\documentclass[10pt,twocolumn,letterpaper]{article}

\usepackage{cvpr}              %

\usepackage[dvipsnames]{xcolor}

\definecolor{cvprblue}{rgb}{0.21,0.49,0.74}
\usepackage[pagebackref,breaklinks,colorlinks,citecolor=cvprblue]{hyperref}

\def\paperID{XXXXX} %
\def\confName{CVPR}
\def\confYear{2024}

\title{Boundless: Generating Photorealistic Synthetic Data for Object Detection in Urban Streetscapes}

\author{
  \begin{tabular}[t]{@{}c@{\hspace{2em}}c@{\hspace{2em}}c@{}}
    \begin{tabular}[t]{c}
      Mehmet Kerem Turkcan \\
      Columbia University \\
      New York, NY \\
      {\tt\small mkt2126@columbia.edu}
    \end{tabular} &
    \begin{tabular}[t]{c}
      Yuyang Li \\
      Columbia University \\
      New York, NY \\
      {\tt\small yl5339@columbia.edu}
    \end{tabular} &
    \begin{tabular}[t]{c}
      Chengbo Zang \\
      Columbia University \\
      New York, NY \\
      {\tt\small cz2678@columbia.edu}
    \end{tabular} \\[2ex]
    \begin{tabular}[t]{c}
      Javad Ghaderi \\
      Columbia University \\
      New York, NY \\
      {\tt\small jg3465@columbia.edu}
    \end{tabular} &
    \begin{tabular}[t]{c}
      Gil Zussman \\
      Columbia University \\
      New York, NY \\
      {\tt\small gil.zussman@columbia.edu}
    \end{tabular} &
    \begin{tabular}[t]{c}
      Zoran Kostic \\
      Columbia University \\
      New York, NY \\
      {\tt\small zk2172@columbia.edu}
    \end{tabular}
  \end{tabular}
}

\begin{document}
\maketitle
\begin{figure*}[t!]
  \centering
  \begin{subfigure}{0.49\linewidth}
    \includegraphics[width=1.0\linewidth]{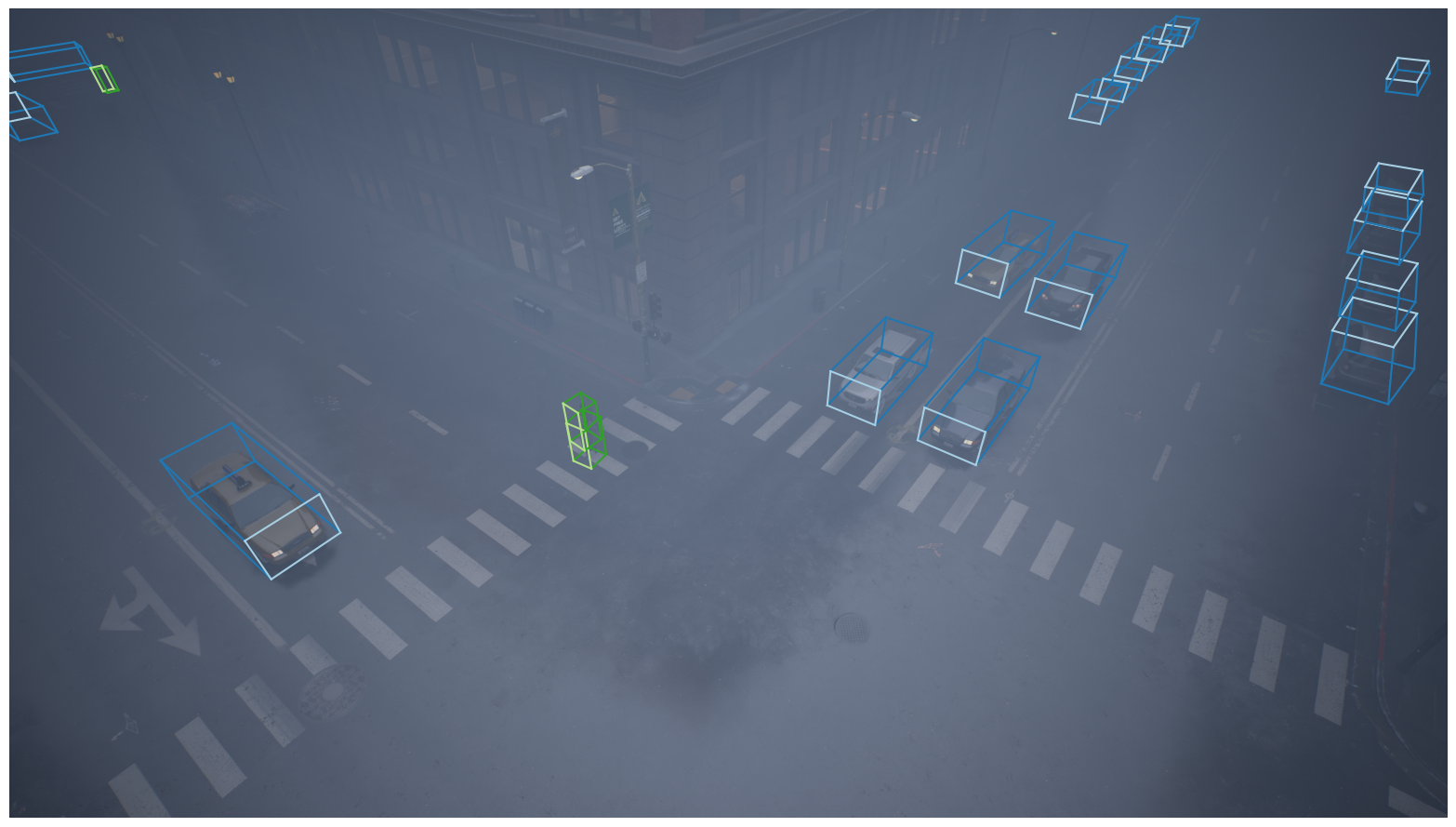}
    \caption{Fog}
    \label{fig:short-c}
  \end{subfigure}
  \hfill
  \begin{subfigure}{0.49\linewidth}
    \includegraphics[width=1.0\linewidth]{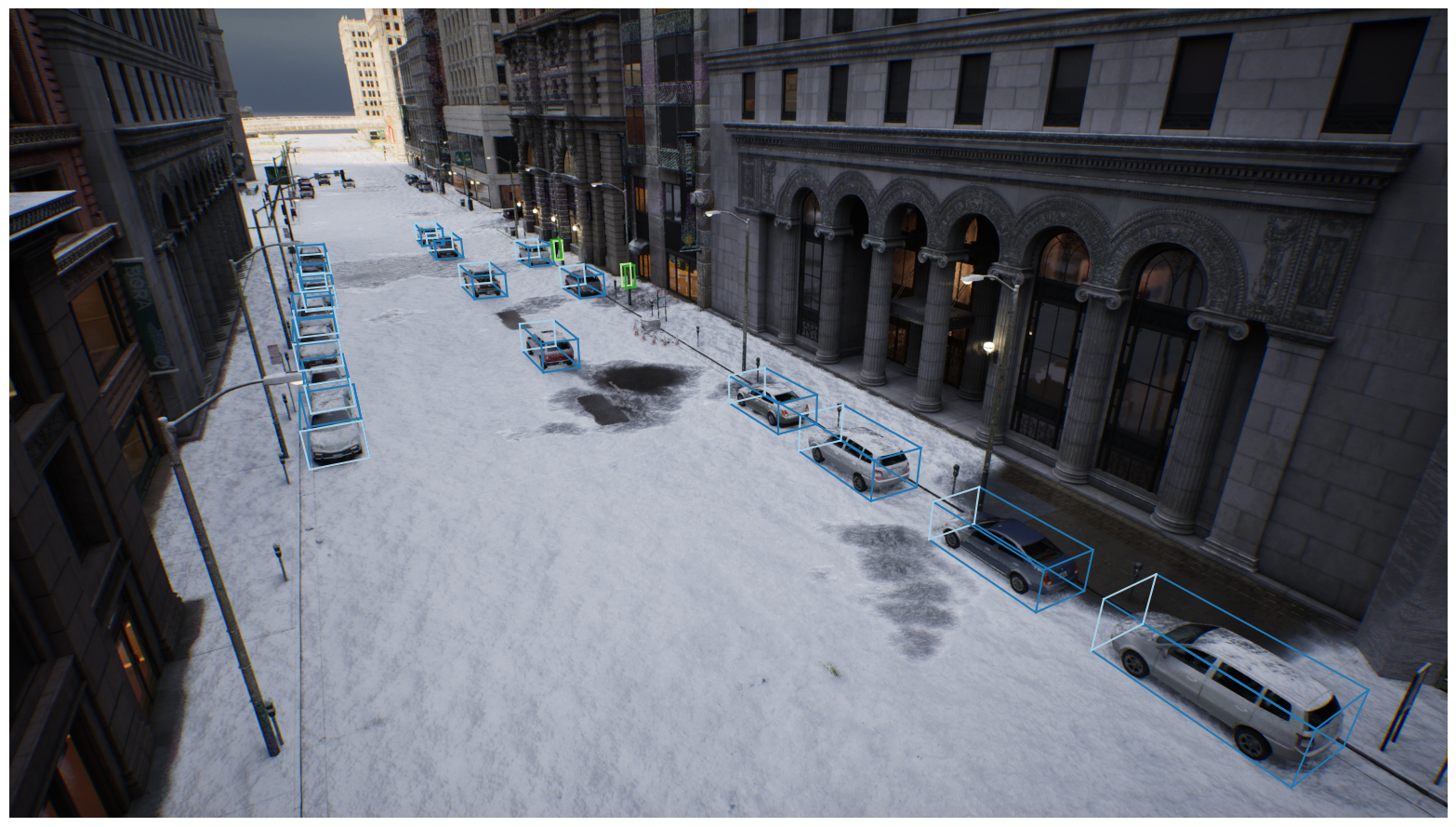}
    \caption{Snow}
    \label{fig:short-a}
  \end{subfigure}\\
  \hfill
  \begin{subfigure}{0.49\linewidth}
    \includegraphics[width=1.0\linewidth]{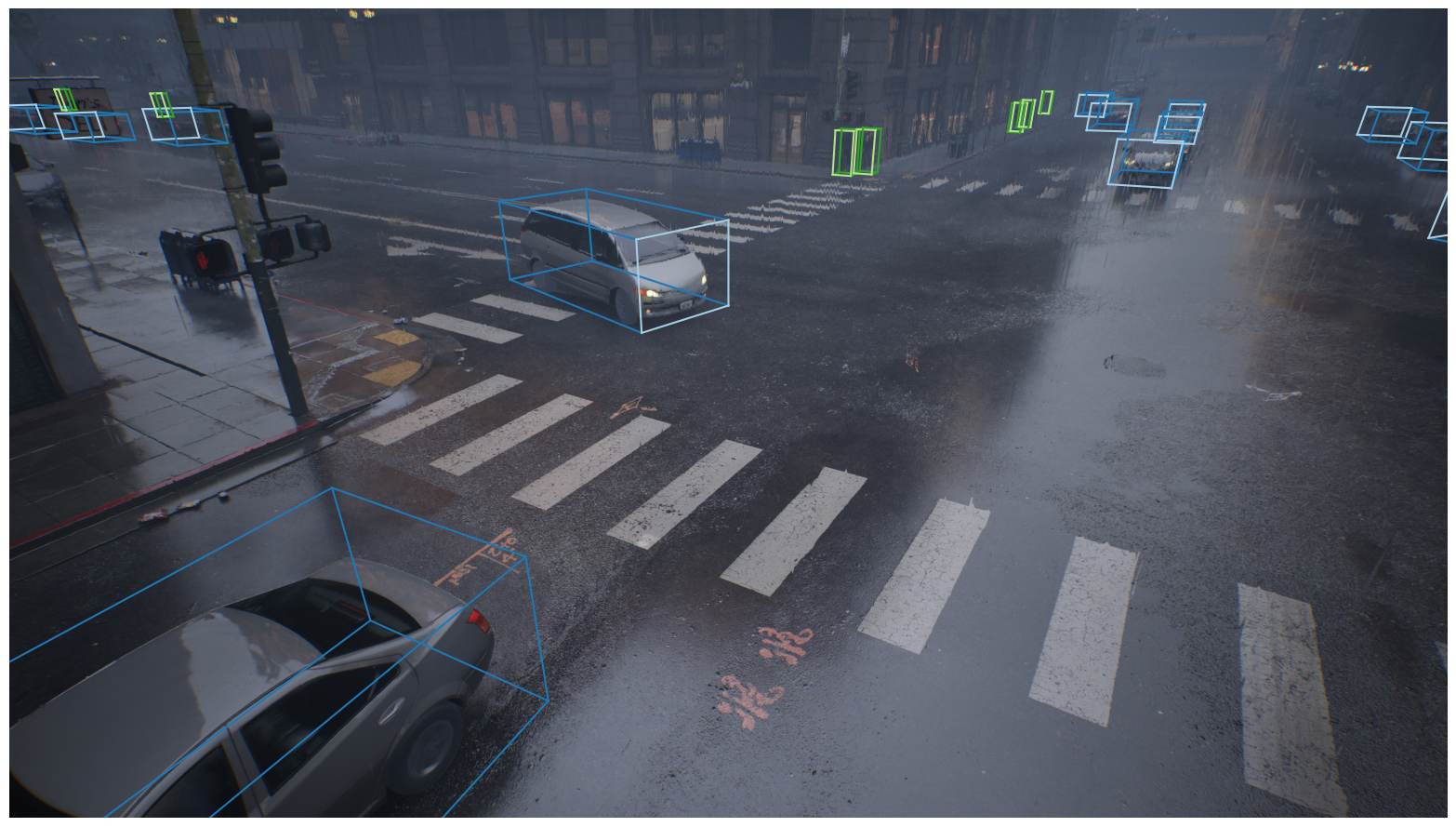}
    \caption{Rain}
    \label{fig:short-b}
  \end{subfigure}
  \hfill
    \begin{subfigure}{0.49\linewidth}
    \includegraphics[width=1.0\linewidth]{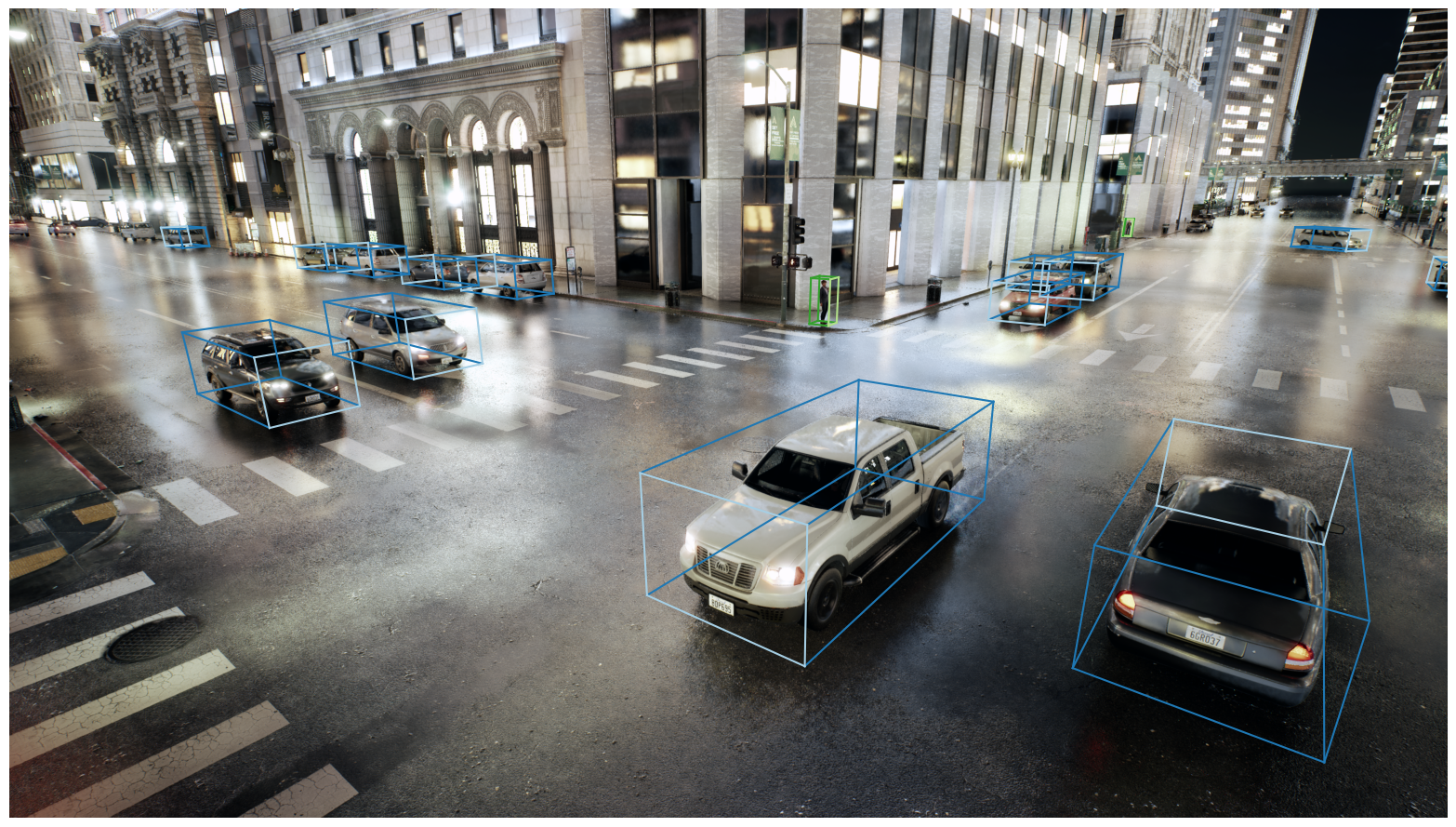}
    \caption{Night}
  \end{subfigure}
  \caption{Boundless enables dynamic weather and environment changes for different situations, with changing time-of-day, weather conditions and background elements for every camera. (a-d) Different weather conditions.}
  \label{fig:boundless_overview}
\end{figure*}

\begin{abstract}
We introduce Boundless, a photo-realistic synthetic data generation system for enabling highly accurate object detection in dense urban streetscapes. Boundless can replace massive real-world data collection and manual ground-truth object annotation (labeling) with an automated and configurable process. Boundless is based on the Unreal Engine 5 (UE5) City Sample project with improvements enabling accurate collection of 3D bounding boxes across different lighting and scene variability conditions. 

We evaluate the performance of object detection models trained on the dataset generated by Boundless when used for inference on a real-world dataset acquired from medium-altitude cameras.
We compare the performance of the Boundless-trained model against the CARLA-trained model and observe an improvement of 7.8 mAP.
The results we achieved support the premise that synthetic data generation is a credible methodology for training/fine-tuning scalable object detection models for urban scenes.

\end{abstract}

\section{Introduction}
\label{sec:intro}
Pedestrian safety and traffic management in bustling cities can be enhanced by using video-based AI systems for real-time monitoring and interaction with street objects \cite{zhang2019edge}. Autonomous vehicles will face challenges due to irregular street layouts, numerous cohabitants, and unpredictable pedestrian behavior \cite{campbell2010autonomous,yaqoob2019autonomous}. This calls for the use of infrastructure-mounted cameras and sensors to gather real-time video data at locations like traffic intersections, where object detection, tracking, trajectory prediction, and high-level reasoning can be performed on edge servers in real time~\cite{kostic2022smart}.

To scale up video monitoring systems in large cities, deep learning (DL) models need to be trained and fine-tuned for hundreds of intersections, each with multiple cameras at varying micro-locations.
Successful training of supervised DL models depends highly on the availability of ground-truth annotated (labeled) data. For traffic intersections, the annotation applies to vehicles, bicycles, pedestrians, and other moving objects, as well as immovable traffic furniture. 

Real-world image collection is complicated by uncontrollable environmental conditions such as variations in lighting, weather, and the unpredictable behavior of transient objects like pedestrians and vehicles.
``Manual'' ground-truth annotation of street objects from arbitrary camera angles requires a large time commitment and monetary resources. 
Data collection in real-world scenarios additionally faces legal issues due to privacy violations. Consequently, existing urban datasets often comprise isolated scenes rather than exhaustive city maps, lacking in the ability to capture the multifaceted nature of cityscapes. These datasets offer limited perspectives, predominantly at eye-level street or high-altitude aerial views, which only partially represent the possible views that one might acquire in the urban deployment of cameras. 

In this work, we investigate the use of synthetic data/image generators that can automatically create ground-truth annotations for training of object detection models, and therefore avoid the complexity and cost of manual ground-truth annotations.

We focus on the generation of synthetic image datasets using Unreal Engine 5\footnote{\url{https://www.unrealengine.com/}} to address the shortcomings of existing object detection datasets in urban environments. We incorporate realistic variable lighting, apply post-processing effects to simulate weather conditions and improve render quality, and make adjustments for the level of detail of graphical assets and accurate capture of bounding boxes. The resulting simulator, which we call ``Boundless'', can simulate diverse conditions, control environmental factors, and automatically generate high-quality ground-truth annotations.  We investigate the suitability of medium-altitude data generated using Boundless for improving object detection performance in a setting for which available real-world training data is scarce.

\section{Related Works}
\label{sec:related_works}

\textbf{Object Detection in Urban Environments.} A large number of datasets focus on low-altitude vehicle and pedestrian detection~\cite{oh2011large,sun2020scalability,personpath22,kovvali2007video,robicquet2016learning,krajewski2018highd,yang2019top,krajewski2020round}. Meanwhile, many other datasets focus on high-altitude aerial environments, where small object detection becomes an important challenge~\cite{zhu2018visdrone,xia2018dota,wang2021tiny,xu2021dot,lai2023stc,lyu2022rtmdet,shermeyer2019effects,chen2020survey}. However, there remains a gap for public mixed-perspective datasets that can adapt to a multitude of different deployment conditions. %

\textbf{Real-Time Object Detection.} Many models have been proposed for object detection. For real-time applications, in recent years single-stage detectors like SSD and YOLO models or transformer-based DETR architecture variants have achieved significant results for real-time detection~\cite{liu2016ssd,redmon2016you,redmon2017yolo9000,redmon2018yolov3,2020arXiv200410934B,wang2022yolov7,yolov8_ultralytics,carion2020end,lv2023detrs}. In this work, we use the state-of-the-art YOLOv8x model for experiments. Rather than model development, our focus is the quality of data used for training. 

\textbf{Synthetic Data Generation.} Realistic 3D simulators have been used extensively for various urban computer vision problems. The SYNTHIA dataset provides a collection of images from a simulated city along with pixel-level semantic annotations, to support semantic segmentation and scene understanding tasks \cite{ros2016synthia}. CARLA, developed in Unreal Engine 4, is an open-source autonomous driving simulator offering extensive resources, including environments and open digital assets explicitly designed for development, testing, and validating self-driving systems \cite{dosovitskiy2017carla}. CARLA-generated imagery has been used for object detection and segmentation \cite{niranjan2021deep,jang2021carfree,lyssenko2021instance}. Extensive work has been conducted on GTA V, using frames collected from this video game for training autonomous driving agents \cite{richter2016playing}. In addition to urban traffic simulators like CARLA, Unreal Engine itself has been considered as a rendering engine for computer vision approaches \cite{agarwal2023simulating, rasmussen2022development, damian2023experimental}. Recently, MatrixCity adapted the photo-realistic City Sample project as a benchmark for training neural rendering models by designing a plugin for saving frames \cite{li2023matrixcity}, focusing on pedestrian- and vehicle-free environments and neural rendering applications. The authors in \cite{hu2023let} used the City Sample project to detect pedestrians and hand-annotated them for this purpose. 

\textit{In contrast to previous work, here we focus on enhancing the available City Sample project to enable accurate and automated data collection for training performant models for urban deep learning applications}. We find that extensive customization is required to enable accurate bounding box annotation collection in Unreal Engine due to a variety of challenges. We further improve the project with live lighting changes and weather and release multiple datasets for medium-altitude object detection, a problem of interest in urban metropolises \cite{duan2021smart}.

\begin{figure*}[h!]
  \centering
  \begin{subfigure}{0.49\linewidth}
    \includegraphics[width=0.5\linewidth]{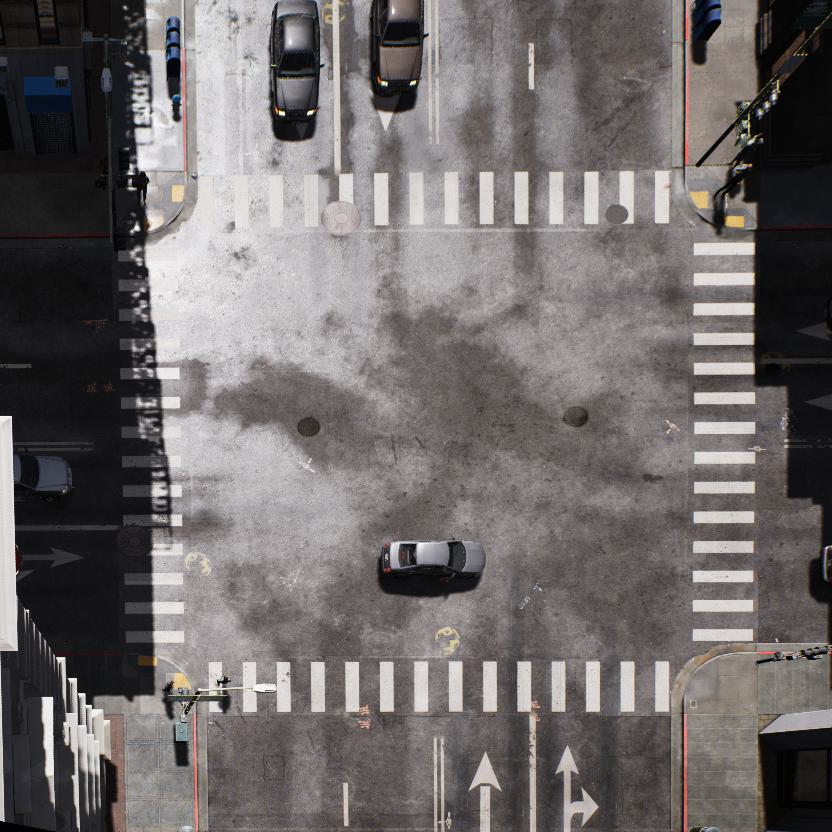}~\includegraphics[width=0.5\linewidth]{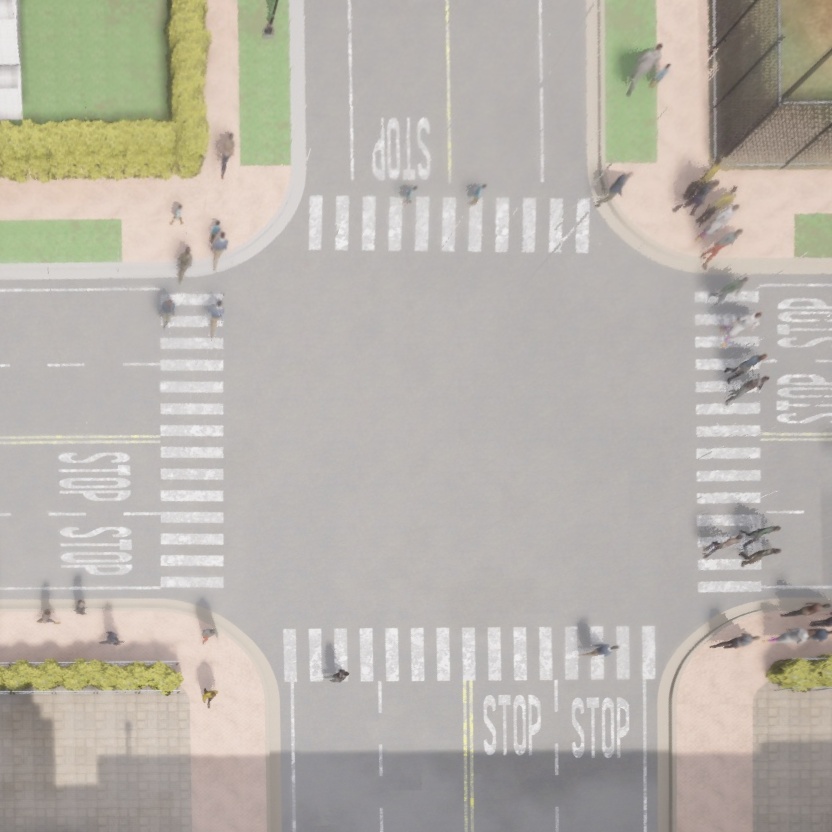}
    \caption{(left) Sample frame of a medium-altitude frame from Boundless; (right) sample frame from CARLA as a comparison against the visual quality of frames rendered using Boundless.}
    \label{fig:short-b}
  \end{subfigure}
  \hfill
  \begin{subfigure}{0.49\linewidth}
    \includegraphics[width=0.5\linewidth]{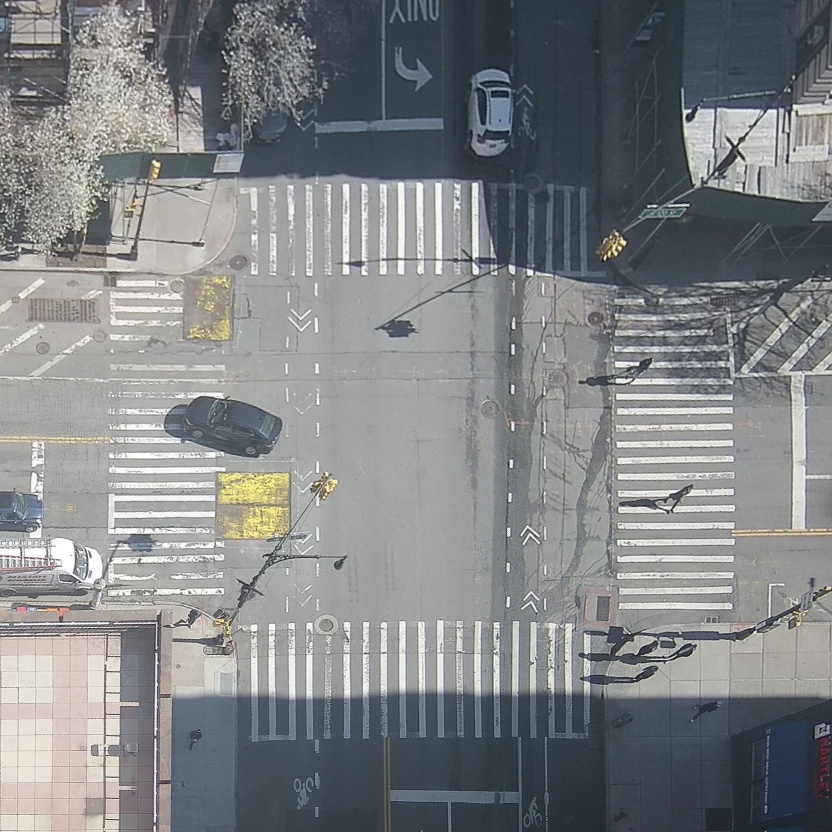}~\includegraphics[width=0.5\linewidth]{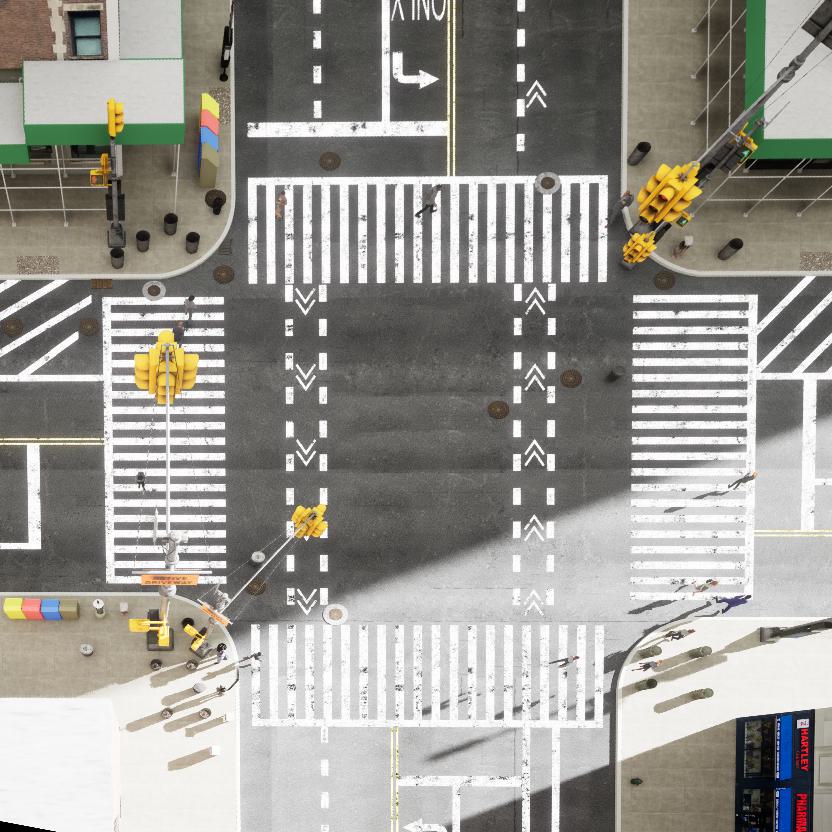}
    
    \caption{(left) An example frame from the real-world validation dataset; (right) sample frame from a digital twin designed to approximate the real-world validation data in Boundless for comparison.}
  \end{subfigure}
  \caption{Examples of frames used for the medium-altitude object detection benchmark.}
  \label{fig:dataset_examples}
\end{figure*}

\begin{table*}[h!]
\centering
\caption{Per-class average precision and mean average precision of YOLOv8x models trained on VisDrone, CARLA and Boundless datasets, evaluated on the real-world validation set.}
\begin{tabular}{|l|l|l|l|} 
\hline
\textbf{Training Dataset} & \textbf{Pedestrian AP@0.5} & \textbf{Vehicle AP@0.5} & \textbf{mAP@0.5}  \\ 
\hline
VisDrone                  & 24.8                       & \textbf{86.9}                    & 55.8              \\ 
\hline \hline
CARLA                     & 14.9                       & 75.4                    & 45.1             \\
\hline
Boundless                     & {24.0}                       & {81.8}                   & {52.9}              \\
\hline
Boundless + Digital Twin                     & {\bf 47.5}                       & {85.7}                   & \bf{66.6}              \\
\hline
\end{tabular}
\label{tab:results_untrained}
\end{table*}

\section{Boundless Simulator Design}
The freely available City Sample project provides an 
environment for North American cityscapes, including vehicles, pedestrians, and traffic lights \footnote{\url{https://www.unrealengine.com/marketplace/en-US/product/city-sample}}%
. To facilitate realistic data collection, we created Boundless by making technical changes to the City Sample project to enable realistic data collection for use in AI applications. We show examples of scenes created with Boundless under different weather conditions, including bounding boxes, in Figure \ref{fig:boundless_overview} to show the capabilities of the simulator. We detail the technical improvements below.

\textbf{Lighting.} 
We allow lighting conditions to be changed dynamically before each new frame collection, thus allowing the scene to change substantially in a single capture session. We implement four new weather conditions corresponding to rain, snow, dust and heat waves. Implemented using decals projected onto the map, the rain and snow effects allow for a more realistic alternative to image-level augmentations \cite{saxena2018automold}. We add particle effects for all these weather conditions. We note that the City Sample comes with a default night-time implementation; however, the stylized and overly dark night weather does not correspond to real-world night-time conditions.

\textbf{Anti-Aliasing.} The default temporal anti-aliasing approach in UE5 produces blurred frames. We replace the approach with MSAA and change the camera settings to output 3840x2160 resolution frames.

\textbf{Level of Detail.} The City Sample project includes three levels of detail for pedestrian and vehicle actors. For the medium and low levels of detail, the substituted meshes have insufficient resolution and detail quality for facilitating real-world applications. We change the available levels of detail for each vehicle and pedestrian agent, enabling the simulator to capture distant object bounding boxes accurately from medium-altitude and street-level scenes.

\textbf{Updated Bounding Boxes.} Due to the design of pedestrian and vehicle actors in the City Sample project which results in large or missing collision boundaries for different 3D meshes, significant changes are needed to correctly capture bounding boxes of objects. We re-compute the level of detail, visibility, and occlusion properties of individual objects in a scene before capturing a frame to make sure annotations of all visible objects are obtained.

\textbf{Export Options.} Boundless exports 3D bounding boxes in the KITTI and 2D bounding boxes in the YOLO format.

\section{Datasets}
{We give an overview of the different datasets we introduce for our experiments in Figure \ref{fig:dataset_examples}. We generated two synthetic datasets using Boundless:}

 \textbf{$\bullet$ Medium-Altitude City Sample Training Set.} The City Sample comes with default city maps. We use Boundless to collect an 8,000-frame dataset from a synthetic intersection from the City Sample project with a static camera angle. All bounding boxes are generated automatically by the simulator. Lighting conditions are changed throughout the training set in between every frame. A frame is saved from the simulation every 3 seconds in simulation time. For comparison purposes, we follow the same approach to create a corresponding dataset consisting of 22,000 frames using the CARLA simulator.

\textbf{$\bullet$ Medium-Altitude Digital Twin Training Set.} We create a realistic 3D digital twin of a real-world intersection within Boundless. We collect 8,700 frames from this highly accurate scene, replicating the camera angles of the medium-altitude real-world validation dataset. 

 In addition to our synthetically generated datasets, we use the following two real-world image datasets:

\textbf{$\bullet$ Medium-Altitude Real World Validation Set.} We create a real-world image dataset collected from a major North American metropolis for one intersection. This validation set consists of 3,084 frames collected on different days with a variety of weather and time-of-day conditions. The dataset contains 12,380 vehicles and 15,225 pedestrian bounding boxes.

\textbf{$\bullet$ VisDrone Dataset.} The VisDrone dataset~\cite{zhu2018visdrone} contains 7,019 images captured from various perspectives, including top-down and ground-level views, with varying camera angles. Although the dataset originally contains multiple object classes, we adapt it to a two-class object detection problem. 
We use the 561-image validation split for reporting our results.

\section{Experiments}
\vspace{-0.05cm}
\textbf{Medium-Altitude Object Detection.} We used a medium-altitude object detection task to demonstrate the capabilities of Boundless, where the amount of data available is scarce compared to other types of urban data. In this task, we seek to detect pedestrians and vehicles from a static camera at $\sim$40m height. We wanted to explore how well a model performs on this task when trained on different datasets. To do so, we fine-tuned a COCO-pretrained YOLOv8x model \cite{lin2014microsoft,yolov8_ultralytics} on three different datasets: (i) VisDrone, (ii) CARLA, and (iii) Boundless. We evaluated the models on a custom dataset collected from a real-world traffic intersection. All models were trained with SGD for 10 epochs at a learning rate of 1e-3.  We show the results of this comparison in Table \ref{tab:results_untrained}. Use of Boundless-generated data using the City Sample map yields a model that greatly outperforms the CARLA-generated data on the real-world validation set, and further exceeds VisDrone performance in this dataset. 

\vspace{-0.05cm}

Motivated by the superior performance obtained using Boundless, we further implemented a digital twin of the real-world intersection from the real-world validation set in the form of a map in Unreal Engine. We collected more data using Boundless from this map to see how much the results could be improved. Using a 3D model as a map, we collect an additional 8,700 frames. Repeating the experiment by adding these additional frames further improves the mAP score significantly, yielding a higher mAP than the use of VisDrone.

\section{Ethical Considerations} 

The real-world dataset collected is deliberately installed at an altitude that makes it impossible to discern the pedestrian faces or to read car license plates. The academic institution at whose facilities the camera is installed provided an IRB waiver for sharing this ``high elevation data" with the general public since the data inherently preserves privacy. 

\vspace{-0.20cm}
\section{Conclusion} 
\vspace{-0.15cm}
We created and described the Boundless simulation platform, and benchmarked it in a real-world medium-altitude object detection task that demonstrates challenges with the deployment of object detection models to urban streetscapes. Our best-performing model, using Boundless and incorporating a digital twin of the real-world testbed, achieved an mAP of 66.6, demonstrating the ability of the simulator to create imagery with sufficient realism to be deployed in real-world scenarios.
With Boundless, we seek to support research on urban object detection for metropolises, where data collection, ground-truth annotation/labeling, and model training face technical and legal challenges.
By releasing the simulator along with collected datasets, we aim to facilitate future research and applications in urban computer vision problems.
\vspace{-0.25cm}
\section*{Dataset and Code Availability} 
\vspace{-0.15cm}
Datasets and code for the experiments in this study are available at \small{\url{https://github.com/zk2172-columbia/boundless}}.
\section*{Acknowledgements} 

This work was supported in part by NSF grants CNS-1827923 and EEC-2133516, NSF grant CNS-2038984 and corresponding support from the Federal Highway Administration (FHA), NSF grant CNS-2148128 and by funds from federal agency and industry partners as specified in the Resilient \& Intelligent NextG Systems (RINGS) program, and ARO grant W911NF2210031.

{
    \small
    \bibliographystyle{ieeenat_fullname}
    \bibliography{main}
}

\end{document}